\documentclass[acmsmall]{acmart}
\AtBeginDocument{%
  \providecommand\BibTeX{{%
    \normalfont B\kern-0.5em{\scshape i\kern-0.25em b}\kern-0.8em\TeX}}}

\setcopyright{acmcopyright}
\copyrightyear{2018}
\acmYear{2018}
\acmDOI{10.1145/1122445.1122456}

\acmJournal{JACM}
\acmVolume{37}
\acmNumber{4}
\acmArticle{111}
\acmMonth{8}

\usepackage{caption}
\usepackage{subcaption}
\usepackage{multirow}
\usepackage{todonotes}
\usepackage{comment}
\newcommand{\red}[1]{\textcolor{black}{#1}}

\usepackage{hyperref}

\begin{document}

\title{Revisiting Contextual Toxicity Detection in Conversations}

\author{Atijit Anuchitanukul}
\authornote{Both authors contributed equally to this research.}

\affiliation{%
  \institution{Contex.ai}
  \city{London}
  \country{UK}
}

\author{Julia Ive}
\authornotemark[1]
\authornote{The author performed the work while at Contex.ai}
\affiliation{%
  \institution{Queen Mary University of London}
  \city{London}
  \country{UK}
}

\author{Lucia Specia}

\affiliation{%
  \institution{Contex.ai}
  \city{London}
  \country{UK}
}


\renewcommand{\shortauthors}{Anonymous}

\begin{abstract}
Understanding toxicity in user conversations is undoubtedly an important problem. Addressing ``covert'' or implicit cases of toxicity is particularly hard and requires context. Very few previous studies have analysed the influence of conversational context in human perception or in automated detection models. We dive deeper into both these directions. We start by analysing existing contextual datasets and \red{find} that toxicity labelling by humans is in general influenced by the conversational structure, polarity and topic of the context. We then propose to bring these findings into computational detection models by introducing and evaluating (a) neural architectures for contextual toxicity detection that are aware of the conversational structure, and (b) data augmentation strategies that can help model contextual toxicity detection. Our results \red{show} the encouraging potential of neural architectures that are aware of the conversation structure. We also \red{demonstrate} that such models can benefit from synthetic data, especially in the social media domain.
\end{abstract}


\maketitle

\section{Introduction}

Understanding toxicity\footnote{We use the term ``toxicity" as an umbrella term to denote a number of variants commonly named in the literature, including content that encompasses hateful, abusive, offensive, aggressive, rude, inappropriate and demeaning behaviours. For the purposes of the experiments, the definition of toxicity may be more specific and is given by the labelling of each dataset.} in user conversations online (e.g., on social networks and other social media platforms) is undoubtedly an important problem both for (a) humans processing such content, either when interacting with others or in the task of content moderation, (b) technological solutions that are aimed at supporting human moderation. As with many other tasks involving human language, the decision on whether a given piece of text is toxic is not trivial. While some occurrences clearly and intentionally use toxic words to abuse or offend, others are more subtle, including the use of sarcasm, references to previous elements in the conversational context or even external elements. 

As it has been argued in previous work, addressing these ``covert'' \red{(implicit)} cases of toxicity requires  context  \cite{jurgens-etal-2019-just,vidgen-etal-2019-challenges,caselli-EtAl:2020:LREC}. Various types  of context have been considered, including conversational and visual context as well as metadata about user features, annotator features or interactions (network information) \cite{ribeiro2018like}.

While we acknowledge the importance of metadata, this information is not always available and also varies from dataset to dataset, depending on the platform used to produce the data. \red{We aim to study platform-agnostic contextual information.} Visual context has been studied in  \cite{gomez2019exploring,yang-etal-2019-exploring-deep,kiela2021hateful} and shown to improve detection results significantly for memes \cite{kiela2021hateful}.
 
In this paper, we focus on textual context, which is prevalent in most social media platforms and can encompass complex linguistic features, making toxicity detection very challenging. Textual context in conversations can include the content of the initial post on which a comment is made, as well as other comments in the conversation thread. 

Very few previous studies have analysed the influence of conversational context in human perception of toxicity in controlled experiments. \citet{pavlopoulos-etal-2020-toxicity} annotated 250 comments on Wikipedia Talk pages in two settings: in isolation and in the presence of the post title and previous comment. They found that 5\% of the 250 samples had their labels flipped, mostly from non-toxic to toxic\red{, when considering the conversational context during annotation}. Similarly, \citet{FBK} annotated 8K tweets from an existing dataset \cite{Antigoni} in isolation and in the presence of 1-5 previous tweets. They found that more context led to 50\% fewer tweets  being considered toxic, and that the longer the context, the higher the chance of a tweet being considered non-toxic. 

While the type and extent of the impact of context differs, both these studies have shown that humans perceive toxicity differently depending on whether they have access to context. 

Both \cite{pavlopoulos-etal-2020-toxicity,FBK} have also used the annotated data to compare prediction models built with and without context. Their contextual models use simple approaches for early or late fusion of context and target comment (i.e., content to be classified), generally by concatenating the segments and giving them as input to the models, or encoding them separately and concatenating the hidden representations. The findings are not very encouraging:  context-aware models are not better than or even harm the performance of context-unaware models. 

We dive deeper into the questions of whether and how context affects human perception and prediction models. When it comes to understanding the impact of toxicity in human perception,  previous  studies have focused on providing a quantification of cases where the presence of context (in {\em phase 2} of the annotation task) has changed human labelling for toxicity (from {\em phase 1} of the annotation task), as well as the direction of change (from toxic to non-toxic or vice-versa). Apart from a few examples of these cases, they have not studied or quantified the reasons why these changes happen. We thoroughly analyse the contextual dataset of tweets released by \citet{FBK} to answer this question. Our observations suggest that labelling errors in {\em phase 1} resolved by the context in {\em phase 2} are mainly changes from toxic {\em phase 1} to non-toxic {\em phase 2}, which were driven by the prevalent positive or neutral polarity of the conversational context.  

Based on these observations, we came to the conclusion that (a) for models/architectural choices, we need to move away from the current approaches which simply perform early fusion of context and target comment (e.g., via concatenation) to take the hierarchy of the utterances in a thread into account; (b) the type of context also plays a crucial role in the prediction models and more instances of cases where the toxicity really matters with diverse polarity and topics are needed. To address these limitations, we study various deep learning architectures that are aware of the conversational structure. In addition, we propose a data augmentation methodology that creates artificial, more diverse context and target utterances in terms of polarity and topics. This methodology uses the state-of-the-art approaches for controlled text generation \cite{gpt-2,style-transformer}. 

Our {\bf main contributions} are thus threefold:
\begin{enumerate}
    \item We provide in-depth analysis of a contextual toxicity detection dataset  \cite{FBK} that reveals the idiosyncrasy of contextual toxicity perception (Section \ref{section:analysis}).
    \item We explore a range of architectures for contextual toxicity detection that are more geared towards conversational context and lead to improved performance (Section \ref{subsec:models}).
    \item We explore data augmentation strategies that help model toxicity (Section \ref{subsec:data-aug}).
\end{enumerate}
We present our experimental setup in Section \ref{section:experimental_settings} and key results and discussion in Sections \ref{section:results} and \ref{section:discussion}.
 
\section{Related Work} \label{section:related-work}

Most research in the area of toxicity detection investigate the use of machine learning algorithms to process individual posts or comments from social media platforms, without any additional context~\cite{zampieri-etal-2020-semeval,waseem-etal-2017-understanding}. While this is a step in the right direction, compared to simple lookup based on keywords (e.g., Hatebase\footnote{\url{https://hatebase.org/}} and The Weaponized Word\footnote{\url{https://weaponizedword.org/}}), this approach is limited to covering ``overt'' cases of toxicity, whereas addressing ``covert'' cases would require  access to more context  \cite{jurgens-etal-2019-just,vidgen-etal-2019-challenges,caselli-EtAl:2020:LREC}. In what follows, we described the few studies in the literature that have attempted to do so for \red{conversational context}, as well as general work in text generation for data augmentation.   

\subsection{Contextual Toxicity Detection}
 
Previous studies on contextual toxicity include those focusing on understanding human perception of toxicity  \cite{pavlopoulos-etal-2020-toxicity,FBK}, creating corpora where annotations are done in context \cite{contextual_abuse,vidgen-etal-2021-introducing,onabola2021hbert,FBK,xu-etal-2021-bot,dialogue_safety,hs-cn-dataset}, and building classifiers where additional context is taken into account, including a single previous comment ~\cite{pavlopoulos-etal-2020-toxicity}, the title of the news article the comment refers to \cite{gao-huang-2017-detecting} or a conversational thread \cite{FBK,dialogue_safety,xu-etal-2021-bot}. 

In terms of human perception, \cite{pavlopoulos-etal-2020-toxicity,FBK} have emphasised the importance of contextual information in understanding the true meaning of comments. For creating computational models that are able to take context into account, the main challenge is the analysis of the conversational structure, which requires modelling long-range dependencies between utterances. This is a general problem in the areas of discourse and dialogue.  Existing work for modelling conversational structure in these areas can be divided in two groups: (a) traditional approaches that summarise conversations into main features (e.g., conversation markers~\cite{niculae-danescu-niculescu-mizil-2016-conversational}), and (b) neural approaches that explore the sequential nature of the conversation by using hierarchical neural network structures~\cite{chang-danescu-niculescu-mizil-2019-trouble}, often equipped with attention mechanisms~\cite{de-kock-vlachos-2021-beg} to focus on the most important parts of inputs at each level. We take inspiration from the latter line of work. 

The works closest to ours are \cite{FBK,dialogue_safety,xu-etal-2021-bot}. As mentioned previously, the study in \cite{FBK} annotated a dataset of tweets in two settings: with and without context. Using the dataset, the study investigated the classification performance of neural and non-neural approaches. For the neural approaches, the study employed a BERT-based model \cite{devlin-etal-2019-bert} which replicates the setup of the Next Sentence Prediction (NSP) task by splitting the last utterance to be classified and its preceding dialogue history into two separate segments. The last hidden state of the model is provided as input into a toxicity classification layer. 

They also experimented with a Bidirectional Long Short-Term Memory (BiLSTM) recurrent network, either encoding the context concatenated together with the tweet as a single text or encoding the context and the tweet separately and then concatenating the two resulting representations. We also propose to encode context and posts separately. However, we do not take the various previous tweets as a single chunk of context, but rather investigate the role of each part of the context.

\cite{dialogue_safety} proposed the {\em Build it, break it, fix it} strategy to create a more robust detection model. The proposed strategy includes humans and models in the loop: models are trained on some initial data, which is incrementally increased with data from adversarial attacks produced by humans to break the current models, an iterative process repeated a few times. The study investigated two different tasks: single-turn task and multi-turn task. The single-turn task models the detection problem with a single utterance, without context. The multi-turn task, which is similar to the scope of our work, considers context (i.e., preceding comments) of each utterance. For that, similar to the work in \cite{FBK}, they use a model with the NSP setup.

The work in \cite{xu-etal-2021-bot} seeks to improve the generation of ``safe'' comments by conversational bots. They also use a human-and-model-in-the-loop strategy where they create additional data by asking crowd-workers to adversarially converse with a chatbot model with the aim of inducing unsafe responses. Using the dataset, a safety classifier network was constructed to detect offensiveness in both the user input and the model output. If both utterances are classified as unsafe, the model would respond with a non-sequitur by randomly selecting a topic from a known list of safe topics. A second approach ``bakes in'' toxicity awareness to the generative model by modifying target responses to incorporate safe responses to offensive input. The first method in this approach was found to filter examples classified as unsafe out of the dataset, consequently causing the model to be unprepared to handle unsafe input at inference time. The second method, which is more robust, replaces the ground truth response of each conversation in the dataset with a non-sequitur if the response or the last utterance in the history is classified as unsafe. 

While created with different aims in mind, the datasets in \cite{dialogue_safety,xu-etal-2021-bot} are very useful by providing conversational structure going beyond one single comment (or post) and the presence of toxic comments. Along with \cite{FBK}, \cite{hateful_qian} and \cite{contextual_abuse}, these make up for six datasets of this nature, all of which we exploit in this paper.

\subsection{Text Generation for Data Augmentation}
 Natural language generation is an NLP area with a range of applications such as dialogue generation, question-answering, machine translation, summarisation, etc. Most recently, different deep learning text generation techniques have been actively applied for data augmentation purposes in the general NLP domain \cite{liu-etal-2020-data,wei-zou-2019-eda}, as well as for specific NLP tasks of machine translation \cite{edunov-etal-2018-understanding, Sennrich2016}, question answering \cite{data-aug-qa}, biomedical NLP \cite{ive-2020}. 

There are many advantages of applying data augmentation techniques in language tasks, as summarised in \cite{data-aug-for-dl}. The overarching goal is to get better performance out of existing  datasets for supervised learning. Data augmentation techniques are also helpful tools to observe model behaviour and exhibit their failures. They also act as a mean of regularisation, which helps mitigate  overfitting. In other words, with data augmentation, models are less prone to learning spurious correlations and memorising unique patterns in the dataset (e.g., numeric patterns in token embeddings). Subsequently, the use of data augmentation leads to better model generalisation.

In the domain of toxicity detection, very few previous studies  investigated the utilisation of data augmentation. The work in \cite{unsupervised-text-style-transfer} proposed an unsupervised text style transfer approach which translates offensive sentences to non-offensive ones. The aim of the study is to encourage users of online social media platforms to change their behaviour of using profanity when posting. That is, when a message to be posted by a user is considered offensive, a polite version of the message is offered to the user. To do so, the proposed approach employed an RNN-based encoder-decoder text generation model and a collaborative CNN-based classifier to provide indirect supervision \cite{unsupervised-text-style-transfer}. During style transfer, the encoder of the generation model encodes an input sentence and its original (i.e., ground-truth) style into a sequence of hidden states. Using the computed hidden states, the decoder receives a target style and outputs a sequence with the desired style. Although this proposed method is effective in detoxifying sequences, we argue that due to the recurrent nature of the text generation model, it is less robust when performing style transfer of long sequences. Thus, we focus our investigation on the state-of-the-art Transformer-based approaches, to be outlined in the proceeding sections.

The study in \cite{counterfactual-data} investigated the effect of using counterfactually augmented data in the robustness of online abusive content detection models. This type of data is achieved by  human-generated instances that are minimally edited to flip their labels (e.g., from positive sentiment to negative sentiment). They conducted experiments on three different constructs: sentiment, sexism and hate speech, using logistic regression and a fine-tuned BERT as classification models. Their findings indicate that the use of augmented data  improves model generalisation to out-of-domain data, justifying the benefit of data augmentation. In addition, the classification models tend to better learn cues and features in the dataset that are highly correlated with the construct of interest. However, generating data manually is time-consuming and costly.

\section{A Closer Look into Human Perception of Toxicity \label{section:analysis}}
We start with an in-depth analysis of the FBK dataset, the contextual dataset of tweets released by \citet{FBK}, where they re-annotated a subset of \citet{Antigoni}'s dataset for which the tweets were still retrievable from Twitter and had at least one previous tweet as context. The~\citet{Antigoni} data was initially created by randomly sampling tweets with subsequent spam filtering during the period of March to April 2017, and annotated independently of context.
The contextual subset of the data was re-annotated via crowdsourcing, using the same group of annotators, 3 months apart, in two conditions: {\em phase 1} -- without context, and {\em phase 2} -- with context, where context can be 1-5 previous tweets. The final label in each condition is the  majority vote amongst three annotators per tweet  (toxic or non-toxic). For our analysis, we selected this data over the other contextual datasets as it offers naturally occurring comments and context which were not sampled by toxic keywords, as opposed to content intentionally created to break models (e.g., \cite{dialogue_safety,xu-etal-2021-bot}).  

The goal of our analysis is to gain insights on why and when human perception changes in the presence of context to better inform  the design of our toxicity detection approaches. First, we observe that, in the original FBK dataset, the aggregated human labels for only 12\% (1070 tweets) of the tweets changed in the presence of context. {\bf 81\% of these changes were from toxic (1) to non-toxic (0)}, with 19\% changing from non-toxic to toxic. \cite{FBK} claims that context helped annotators understand cases of toxic words used in sarcastic or ironic ways, or unclear cases due to references to other tweets, but they have not provided any details on this. We further analysed these changes in all 1070 samples from phase 1 to phase 2 (\textit{1 $\rightarrow$ 0} or \textit{0 $\rightarrow$ 1} flips) and provide a broader categorisation of the reasons. The annotation was performed by one of the authors in this study, who is a fluent speaker of English. We have identified the following categories as reasons for changes in labels from phase 1 to phase 2:
\begin{itemize}
    \item {\bf C1}: The context was indeed needed to determine that a comment is toxic, as stated in \cite{FBK};
    \item  {\bf C2}: Annotators made an incorrect judgement in phase 1, the comment should have been deemed toxic/non-toxic regardless of the context;
    \item {\bf C3}: Annotators made an incorrect judgement in phase 2, potentially misled by the (lack of) toxicity of the context; the comment should have been deemed toxic/non-toxic regardless of the context; 
    \item {\bf C4}: The comment is unclear, with or without context, so the annotators' choice was arbitrary. 
\end{itemize}

Table \ref{font-table} shows the results of our annotations. We have noticed that in 55\% of the analysed cases the context was indeed helpful. In 14\% of the cases, the toxicity label has been deemed arbitrary. These are generally cases with short context referring to events outside of the conversation which cannot be inferred from it. In 31\% of the cases, we did not observe any particular importance of the context for the perception of toxicity. In Table \ref{font-table} we show examples of each category, where we use P\{x\} to indicate the users involved in each conversation. In all of our experiments on the FBK dataset (see~Section~\ref{section:experimental_settings}), we removed examples where the annotator choice was deemed arbitrary (14\%, 155 C4 examples), and we corrected the labels of the examples in C3 (10\%, 106 examples) in phase 2. We note that cases in category C2 were already corrected in the FBK annotation.  

\begin{table*}[]
\begin{footnotesize}
\begin{tabular}{l|p{4.5cm}|p{4.5cm}|c|c}
\toprule \textbf{Category} & \textbf{Target} & \textbf{Context} & {\bf I} & {\bf C} \\ \midrule
C1 (55\%) & P1: I love toaster strudels. But this one day I was at my aunts and she asked if I wanted any. OF COURSE & P1: took a bite out that hoe and eggs and cheese came out [Loudly Crying] the disrespect. I was like wtf is this!? [Nauseated Face] & 1 & 0 \\ \hline
C2 (21\%) &  P1: That Trump monkey wannabe definitely wants to boot u into a show trial. It looks grim. U need a drone w camera 2 rescue u out  & P2: Plans to arrest me sometime after Sunday in violation of political asylum law [link] … Background [link] & 0 & 1 \\ \hline
C3 (10\%) & P1: Me either, lorbe. It's one of the reasons I can't stand to watch him or listen to him. He's a disgusting excuse for a human. & P2: Why does it seem like everyone has forgotten about the fact that we have a President who sexually assaults women? [TAG1]  [TAG2] P3: [Happy Person Raising One Hand] haven't forgotten & 1 & 0 \\ \hline
C4 (14\%) & P1: I'm chill u motherfucker. Im fucking relaxed & P1: What do u see me as & 1 & 0 \\
\bottomrule
\end{tabular}
\end{footnotesize}
\caption{\label{font-table} Proportion of samples and examples of conversations between different users in the four categories of changes in labels from phase 1 to phase 2. `Target'' is the tweet being annotated,  ``Context'',  the previous tweet(s), ``I'' the annotation in isolation, ``C'' the contextual one: toxic (1) or non-toxic (0). P\{x\} indicates the users involved in each conversation.}
\end{table*}

We also analyse the reasons why the context influences the toxicity perception for a random 250-sample subset of cases in C1, C2 and C3 (186 non-toxic and 64 toxic examples). The results show that the \textit{1 $\rightarrow$ 0} flips happen mostly with the neutral context (69\% of the examples for this flip type). Such examples tend to contain a lot of profane words that actually have no toxic intention but are rather used as intensifiers \red{(see example in row 1 of Table~\ref{reasons-table})}. The \textit{0 $\rightarrow$ 1} flips happen mostly because of the influence of negative context (42\% of the examples for this flip type). Toxic examples where the context is crucial to detect sarcasm make around 13\% of the toxic annotated examples. Such cases can be very difficult even for humans. Examples of our annotations are given in Table~\ref{reasons-table}. Our observations confirm that the sentiment of the conversation plays a crucial role to determine the comment's toxicity. Neutral and positive topics usually concern discussions on personal matters: family, food, hobbies, etc. or general positive content (admiration, excitement, etc.) (see examples in rows 1 and 4 in Table~\ref{reasons-table}). Negative context or context leading to sarcasm are usually context discussing politics (see examples in rows 2 and 5). Ambiguous cases rely on additional information to clarify the meaning of the comment (see example in row 6). Sometimes toxicity is not coming from the author of the post but they repeat words of somebody else (see example in row 7). We also need context to understand if this is the case. Another important observation is that the conversational structure of tweets is very irregular and each new post may refer to any previous posts in the thread (see examples in rows 2 and 3). Furthermore, with potentially a large number of participants, tracking the exchanges between them may be very difficult. Hence, it is crucial to be able to model this structure in flexible way with appropriate architectures.

\begin{table*}[]
\begin{footnotesize}
\begin{tabular}{p{0.05cm}|l|p{4.5cm}|p{4.5cm}|p{0.05cm}|p{0.05cm}}
\toprule
& \textbf{Category} & \textbf{Target} & \textbf{Context} & {\bf I} & {\bf C} \\ \midrule
1 & Neutral context (58\%) &  P1: snakes the devil, that's why I like to see mongooses \textbf{fuckin em up}. slithery \textbf{sinister bastards}  & P2: A missing man was discovered inside a 23-foot python. [LINK] … & 1 & 0 \\ \hline
2 & Negative context (16\%) & P1: use that \textbf{ugly ass} design instead of the oufit designed by kishimoto & P1: Y'all really think \newline P2: is not going to sexualize sarada the way ikemoto pedophile ass has been doing yet why else would they~ & 0 & 1 \\ \hline
3 & NA (13\%) & P1: If u want we can do some \textbf{fucked up} things if u want :)  & P1: Hey bro, im the guy darkpetal - maybe we can do some other stuff ;) \newline P2: The hacker Guy? \newline P1: Haha i usually dont kill people ;D but meh this was my fun day on gta & 1 & 0 \\ \hline
4 & Positive context (6\%) & P1: I'm gonna \@ because you are 100000\% correct the bulls are going to the finals \textbf{Fuck} the Cleveland Cavaliers! Go Bulls!'!!!! & P2: Bulls the best team in the NBA, don't @ me & 1 & 0 \\ \hline
5 & Sarcasm (4\%) & P1: This is adorable. You can't answer direct questions. Not a democrat but nice overreach. & P1: Ask your father he would know. \newline P1: It's adorable that you still provide him a reacharound even after death. Very committed you are. \newline P1: I accept your concession snowflake. \newline P1: Poor kid haz a sad. You do know though that you posted a forgery? \newline P1: Trickered? Do you even know what "triggered" even means or did you just hear it online and decided to be a follower and repeat it? & 0 & 1 \\ \hline
6 & Ambiguity (2\%) & P1: turns out he's just got out of prison for beating his ex girlfriend, I'm \textbf{fucking} disgusted and disappointed in my mum & P1: Okay so I thought there was something weird about my mums new boyfriend,I was getting bad vibes and he was saying gross things about women & 1 & 0 \\ \hline
7 & Citation (1\%) & P1: he's said who are these people cuz I wanna sue them. I \textbf{fucking} fell out!!!!  & P1: The turbo tax Humpty commercial is funny [Face with Tears of Joy] \newline P2: favorite commercial right now lol  & 1 & 0 \\ 
\bottomrule
\end{tabular}
\end{footnotesize}
\caption{\label{reasons-table} Examples of the various categories of changes related to the presence of context. `Target'' is the tweet being annotated,  ``Context'',  the previous tweet(s), ``I'' the annotation in isolation, ``C'' the contextual one: toxic (1) or non-toxic (0). P\{x\} indicates the number of users involved in each the conversation. Reason N/A was assigned when the reason for the toxicity flip was not possible to establish. Bold highlights profanity words in classified posts.}
\end{table*}

Our analysis and findings emphasise the need for context, as well as the complexity of interpreting a comment for toxicity, even in the presence of context. This motivated the two directions we pursued in the remainder of this paper: (a) better contextual detection models that adequately explore the structure of the conversation (rather than simply concatenating target comments and context) and \red{(b) data augmentation strategies to  create more diverse and relevant examples where context matters.} We present our approaches for these two directions in the next section.

\section{Methodology} \label{section:methodology}

In this section, we present our context-aware approach for toxicity detection, followed by the data generation strategies we employ to make the context more informative.   

\subsection{Context-aware Models}\label{subsec:models}

\begin{figure}[!h]
\begin{subfigure}{0.33\textwidth}
  \includegraphics[width=0.9\linewidth]{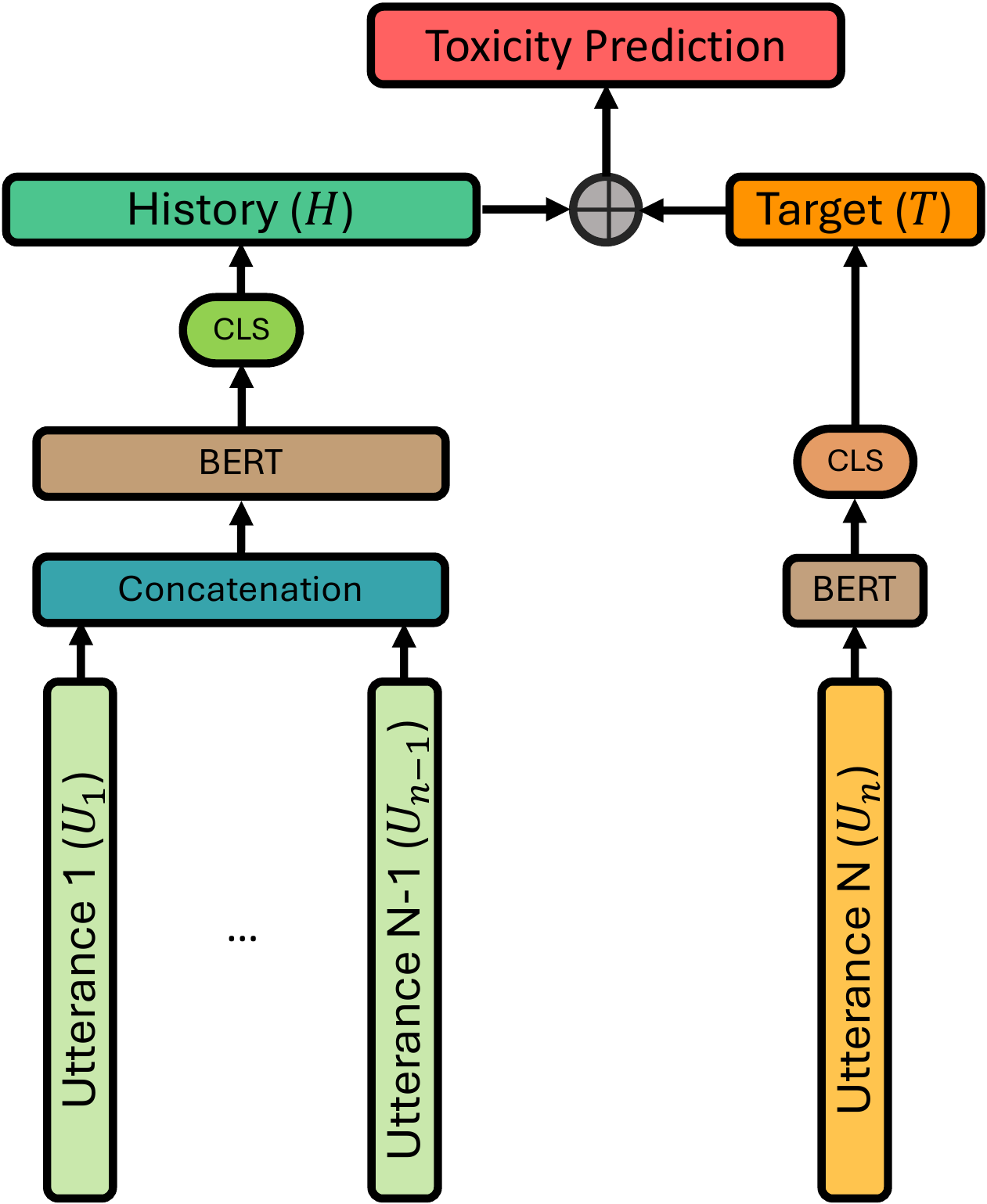}
  \caption{ContextSingle}
  \label{fig:contextsingle}
\end{subfigure}%
\begin{subfigure}{0.33\textwidth}
  \includegraphics[width=0.9\linewidth]{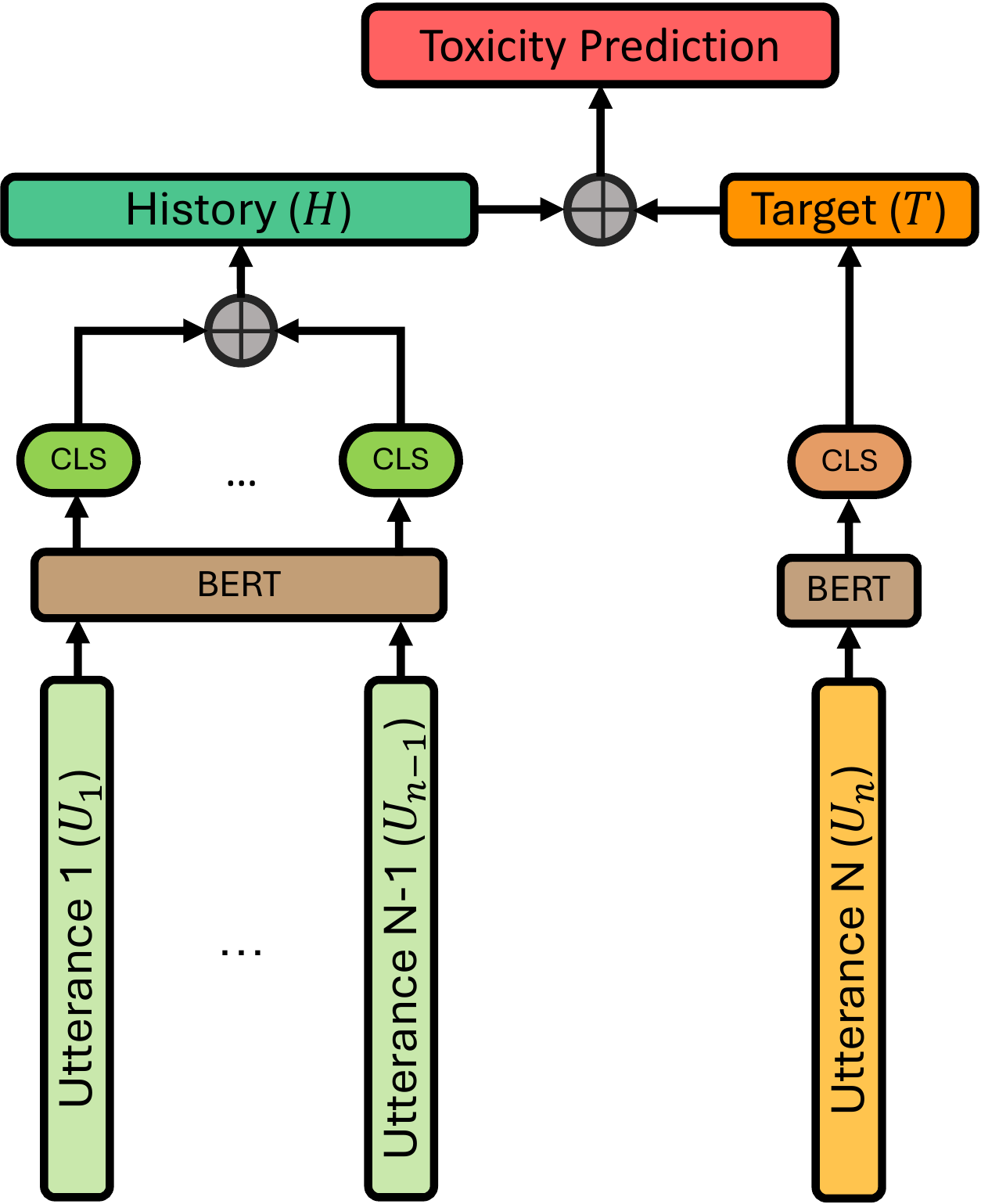}
  \caption{ContextSum}
  \label{fig:contextsum}
\end{subfigure}
\begin{subfigure}{0.33\textwidth}
  \includegraphics[width=0.9\linewidth]{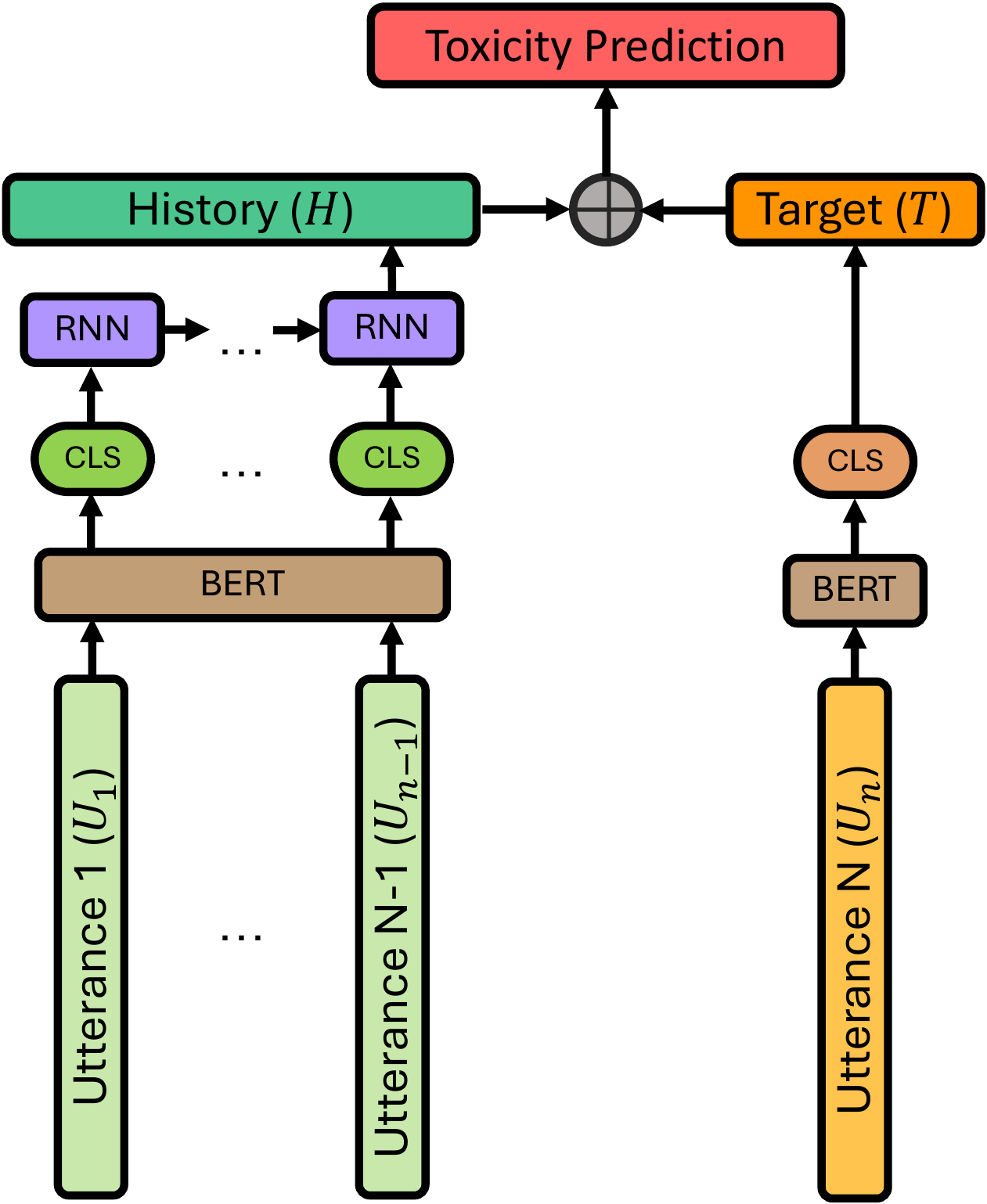}
  \caption{ContextRNN}
  \label{fig:contextrnn}
\end{subfigure}
\caption{Our ContextSingle, ContextSum and ContextRNN architectures. Each utterance $U$ is represented by its BERT CLS token embedding. The context history ($H$) representation is taken as the summary CLS BERT representation for the concatenated context utterances (ContextSingle), the sum of the context representations (ContextSum) or the last hidden state representation from a GRU model (ContextRNN). The history representation is concatenated with the representation of the target post ($U_n$) to produce toxicity predictions.}
\label{fig:context-architecture}
\end{figure}

To better handle the conversational structure, we implement and experiment with three architectures that encode the context separately from the target comment, using late and model fusion techniques to combine them. Specifically, as shown in Figure~\ref{fig:context-architecture}, we encode the context as a sequence of sentences ($U_{1} \dots U_{n-1}$), each represented by its BERT sentence (CLS) embedding. Those embeddings are then summarised into a history representation ($H$). We investigate three summary alternatives, namely taking the summary token from BERT for the concatenated context utterances ({\bf ContextSingle}), the sum of the context representations ({\bf ContextSum}) or the last hidden state representation from a GRU model ({\bf ContextRNN}). The history representation is then fused with the representation of the target post ($U_n$) by concatenation, and this is fed into a final classification layer followed by a sigmoid activation to  classify the sample as toxic or non-toxic. 

\subsection{Data Augmentation}\label{subsec:data-aug}
Given the cost of annotating toxicity in context, most datasets are relatively small. Here we take FBK as a representative example of such small datasets and  explore data augmentation strategies to improve  prediction performance.
More specifically, we explore two data augmentation approaches for both the target and its context: a \textit{generative} approach and a \textit{transformative} approach. The former generates new content based on the original distribution in the training data, while the latter modifies existing training data, creating a variant of the original training data. To implement these approaches, we explore (a) a fine-tuned GPT-2 \cite{gpt-2} language model and (b) the Style Transformer model \cite{style-transformer}. 

GPT-2 is a large state-of-the-art Transformer-based language model \cite{gpt-2}, pre-trained on over 8 million documents. During the pre-training phase, language models such as GPT-2 are known to capture general knowledge about the language. They then need to be adapted to the task at hand by fine-tuning, using labelled data that reflect the expected distribution. We create fine-tuned models to either generate toxic data or non-toxic data so that the text generated would reflect both the style and the content of the fine-tuning dataset. 

For inference, language models require a prompt to initiate the generation as a continuation of this prompt. They then produce one word at a time, conditioned on the previously generated words and the prompt. We consider two types of prompts: the original context to condition the generation of the target comment, and the target comment to condition the generation of context. Note that this last setup breaks the sequential nature of the conversation, nevertheless it might still be suitable for the type of data we used (tweets), given that the chronological sequence of messages does not necessarily indicate a discursive sequence.

Style Transformer~\cite{style-transformer} is a model with an auto-encoder architecture that learns to disentangle the style from the content of the input text, learning separate representations for style and content. The goal is to rewrite (transform) sentences  with a desired style while preserving the content from the original sentence. This is done by learning the network to separate and reconstruct from its own output the original input text and its style (see details in Appendix~\ref{app:styletrans}). For inference, the model takes the tokens of the original text as its input and attempts to rewrite them given the requested style. The model thus has access to the entire input text during the generation. We use Style Transformer to re-write target comments or context into their opposite styles (toxic or non-toxic). It is important to note that we train the Style Transformer model from the start without using a pre-trained model.\footnote{\red{As the Style Transformer is not designed to be used for pre-training, we had to train it from scratch.}}

\section{Experimental Settings} \label{section:experimental_settings}

\subsection{Datasets}

In this section, we describe all datasets used in this paper, which come from various sources and were sampled and labelled using different strategies. In all datasets, we consider binary labels distinguishing toxic cases from the non-toxic ones. In all cases, we have applied the pre-processing algorithm for the social media text as provided by the Ekphrasis library~\cite{baziotis-pelekis-doulkeridis:2017:SemEval2}. It pre-processes text to remove URL addresses and usernames, as well as replace emoticons with their corresponding textual expressions. 

\paragraph{\bf FBK} The FBK dataset \cite{FBK}, which we used in the analysis in Section \ref{section:analysis}, is a subset of the dataset for abusive language detection, introduced in \cite{Antigoni}. The original dataset was formed by sampling tweets using the Twitter API. Using provided tweet IDs, the study in \cite{FBK} queried Twitter API to extract context for tweets (i.e., preceding tweets in the same thread). After filtering out the tweets without context, the dataset was annotated in two steps: the first one without context and the second one with context. As mentioned in Section \ref{section:analysis}, we use the revised version of this dataset (including the removal of cases in C4 and the re-annotation of cases in C3) in all our experiments. This revised version is relatively small, with a total of 7863 samples. We have created our own data split to make sure samples where context is important are in all  splits. We refer to this full dataset as {\em full}. We also created a much smaller sample containing only cases where the toxicity labels changed from phase 1 to phase 2. We refer to this set as {\em flipped} (809 samples).  Instances in these variants of the FBK dataset contain between 1 and 5 tweets.

\paragraph{\bf Build-It Break-It Fix-It (BBF)}
The BBF dataset from \cite{dialogue_safety} was constructed by collecting utterances produced by crowd-workers. They were instructed to intentionally continue segments of existing conversations by producing offensive utterances that the classifiers developed in the study would incorrectly label as safe (non-toxic). The dataset therefore contains contextual information of each produced utterance in the general conversational domain. In terms of annotation, adversarial utterances produced by the crowd-workers were annotated as unsafe. To combine this data with safe examples, dialogue examples from ConvAI2 \cite{convai2} which are labelled as safe by two crowd-workers were added. Instances in the dataset contain between 3 and 6 utterances.

\paragraph{\bf Bot-Adversarial Dialogue (BAD)} The BAD dataset~\cite{xu-etal-2021-bot} was formed by collecting conversations between crowd-workers and the dialogue models (i.e., chatbots). Crowd-workers were instructed to adversarially converse with the bots to instigate the generation of unsafe (toxic) messages by the bots. Each conversation contains between 6 and 32 utterances. Each conversation is then broken down in various numbers of samples with each conversation turn forming a separate training example. That is, one sample contains one utterance of the dialogue along with 0 to 31 preceding utterances, as the two examples shown in Table \ref{Tab:BAD_examples}. This data is also in the general conversational domain. Responses generated by the bots were annotated by 4 crowd-workers, including the person conversing with them, while human-generated utterances were labelled by 3 crowd-workers. An utterance was considered to be unsafe if at least one of the following categories was tagged: Hate Speech, Personal Attack, Profanity and Other Offensiveness.

\paragraph{\bf Hateful Qian Reddit (HQR) and Gab (HQG)}  The HQR and HQG datasets \cite{hateful_qian} were constructed by collecting conversational data from Reddit and Gab, respectively. The data collection procedure followed a keyword-based strategy. That is, a list of hate keywords was used to first identify potentially hateful comments. Then, the conversational context of each comment was reconstructed by considering all the preceding and proceeding comments in the thread. Each extracted comment was then annotated by crowd-workers whether it contains hate speech or not. Similar to the BAD dataset, we break down each conversation thread to multiple samples with each conversation turn forming a training sample. Instances in both datasets contain between 1 and 20 comments.

\paragraph{\bf Contextual Abuse Dataset (CAD)} The CAD dataset \cite{contextual_abuse} was built by sampling posts from Reddit, specifically in subreddits which potentially have higher-than-average levels of abuse. Annotation was carried out by crowd-workers for each comment or title of the post. The crowd-workers were instructed to classify each comment to one or more of these primary categories: Identity-directed abuse, Affiliation-directed abuse, Person-directed abuse, Counter speech, Non-hateful slurs and Neutral. Additionally, the annotators were asked to specify, for toxic cases, whether they require context when annotating each comment. In our experiment, we treat all the three abuse-related categories as toxic and extract context for each comment as all the preceding comments or titles of the same post. We refer to the full dataset as {\em CAD full}. We also created a subset of this dataset with only instances where annotators indicated that context was needed for the decision on the label to be toxic. We refer to this version as {\em CAD context}. Instances in the dataset contain between 1 and 3 post comments or titles.

\begin{table*}[]
\begin{footnotesize}
\begin{tabular}{p{0.05cm}|p{4.5cm}|p{4.5cm}|c}
\toprule 
& \textbf{Target} & \textbf{Context} & \textbf{Gold} \\ \midrule
1 & Bot: That's nice. What do you do for fun? I like to read the newspaper. & P1: a new family just moved into my neighborhood & 0\\ \hline
2 & 
 P1: the family that moved in is making lots of noise and being rude, acting like animals  & P1: a new family just moved into my neighborhood \newline
 Bot: That's nice. What do you do for fun? I like to read the newspaper. & 1\\ 
\bottomrule
\end{tabular}
\end{footnotesize}
\caption{Examples from the BAD dataset showing how a dialogue instance between a crowd-worker and a bot is broken down to multiple instances, where 1 indicates a toxic comment and 0, non-toxic.}
\label{Tab:BAD_examples}
\end{table*}

Table~\ref{Tab:dataset_stats} summarises the statistics of all our datasets, using the provided splits for BBF, BAD and CAD and our custom splits for FBK, HQR and HQG. Note that the FBK, BBF and CAD datasets are heavily imbalanced, while BAD, HQR and HQG are relatively more balanced.

\begin{footnotesize}
\begin{table}[h!]
   \centering
   \begin{tabular}{l|c|c|c|c}
        \toprule
        {\bf Dataset} & {\bf Train/Valid/Test/Total} & {\bf Toxic} &  {\bf Context Len} & {\bf Target Len}\\
        \midrule
        FBK full & 6316/591/956/7863 & 9.6\%/10.3\%/15.8\%/10.4\% & 49.7 & 20.2\\
        FBK flipped & 329/85/395/809 & 18.0\%/17.6\%/16.5\%/17.2\% & 44.1 & 17.0\\
        BBF  & 20524/2480/2472/25476 & 11.7\%/12.1\%/12.1\%/11.8\% & 32.1 & 10.1\\
        BAD  & 69274/7002/2598/78874 & 39.3\%/39.5\%/36.3\%/39.2\% & 95.6 & 15.2\\
        HQR  & 17124/2255/2331/21710 & 24.4\%/23.7\%/22.1\%/24.1\% & 140.7 & 47.8\\
        HQG  & 25638/3214/3187/32039 & 44.1\%/43.6\%/45.4\%/44.2\% & 63.3 & 27.7\\
        CAD full & 13584/4526/5307/23417 & 18.5\%/18.4\%/18.2\%/18.4\% & 49.5 & 30.8\\
        CAD context  & 745/309/356/1410 & 100.0\%/100.0\%/100.0\%/100.0\% & 39.5 & 25.5\\
        \bottomrule
    \end{tabular}
  \captionof{table}{Dataset statistics: number of samples in each split, proportion of toxic cases, average lengths (in words) of context and target comment.}
  \label{Tab:dataset_stats}
\end{table}
\end{footnotesize}

\subsection{Baselines} 

Following the best practices in the domain~\cite{xu-etal-2021-bot}, we build several baseline models based on the same pre-trained BERT as in the previous section.

\begin{itemize}
\item \textbf{TargetOnly} is a BERT-based model that takes only the target comment as input. 
\item \textbf{TextConcat} is the same BERT-based model that takes in the target comment concatenated together with the context utterances in the reverse chronological order, ensuring that the last context utterances are not cut off by the maximum sequence length. This is the most common early-fusion approach used in previous work.
\item \textbf{NextSentPred} is our re-implementation of the models of~\cite{xu-etal-2021-bot} and~\cite{FBK} where the concatenated context is input into the BERT model as the previous sentence and the target sentence is input into the BERT model as the next sentence mimicking the BERT next sentence prediction (NSP) setup. The model is then fine-tuned for the binary toxicity prediction using the resulting BERT CLS token embedding.
\end{itemize}

For a broader perspective, we also compare to three online tools for toxicity detection: (1) the Perspective API,\footnote{\url{https://www.perspectiveapi.com/}} (2) the Azure Content Moderation,\footnote{\url{https://azure.microsoft.com/en-gb/services/cognitive-services/content-moderator/}} and  (3) Clarifai.\footnote{\url{https://www.clarifai.com/use-cases/content-moderation/}} We feed into those tools only the text of the target comment (TargetOnly) as they are non-contextual in nature. These tools output a probability distribution for more fine-grained types of toxicity. To make them into binary labels, after obtaining classification results, we compute the final toxicity scores as follows. For Perspective, we take the predicted probabilities in the \texttt{toxicity} class as the toxicity scores, since this is a general class that includes all types of toxicity. For Azure, we define the toxicity score as the maximum  probability of all three predicted categories: sexually explicit or adult language, sexually suggestive or mature language, and offensive language. For Clarifai, we take the predicted probabilities in the \texttt{toxic} class. 

Additionally, to assess the importance of context in each dataset, we perform a lexicon-based classification on the text of the target comment (TargetOnly) using the English lexicon of The Weaponized Word\footnote{\url{https://weaponizedword.org/lexicons}} (WW). 

\subsection{Setups for Data Augmentation}
We focus our data augmentation experiments on the FBK dataset due to (a) its extreme label imbalance, (b) its small size and (c) the poorer performance across all of our models compared to their performance on other datasets.

\subsubsection{Synthesising Toxic Examples\label{ssec:data_aug}}

To address the label imbalance, we generate toxic examples conditioned on the context sampled from the training dataset. We fine-tune a GPT-2 generation model \cite{gpt-2} using only the toxic part of the data. We augment toxic samples by generating toxic target comments conditioned on the sampled non-toxic context utterances. This is to avoid the bias with the data used to fine-tune the language model. We take the number of the generated toxic instances necessary to reach the balance (50-50\%) of toxic and non-toxic examples in the final training data.

\subsubsection{Synthesising Adversarial Examples}
To address the issue of data sparsity where the context matters, we design the following adversarial data augmentation procedure:
\begin{itemize}
    \item \textbf{1-0 Flip Target} - given the context (preceding non-toxic examples), we generate target tweets containing toxic (mainly profanity) words and mark this data as \textit{non-toxic}. We generate 5K such examples to keep the original number of non-toxic examples in the~\cite{FBK} non-flipped training data (Table~\ref{Tab:dataset_stats}). See example 1 in Table~\ref{tab:gen-examples-trg}.
    \item \textbf{0-1 Flip Target} - given the context (preceding toxic examples), we generate target tweets without explicit toxicity words and label this data as \textit{toxic}. We generate 500 such examples to match the original data label distribution. See example 2 in Table~\ref{tab:gen-examples-trg}.
    \item \textbf{Flip Context} - given the non-toxic tweet, we generate \textit{toxic} context and label this data as \textit{toxic}. We generate 5K such examples to augment the number of toxic examples in the \cite{FBK} data. See example in Table~\ref{tab:gen-examples-context}.
    \end{itemize}

With any of these strategies there is no guarantee that the labels will be correct, however, we expect most of the data will be correctly labelled and the models will be robust enough to deal with some amount of noise.

All the setups are replicated for both the generative (GPT-2) and transformative (Style Transformer) generation approach. Note that transformative examples in Tables~\ref{tab:gen-examples-trg} and~\ref{tab:gen-examples-context} are unsurprisingly very close to the original data. Within each approach, we then experiment with different combinations of data types, with or without adding the real training data. We carry out our experiments by augmenting the training dataset of ContextSingle, our best contextual model on the FBK dataset.

\begin{table*}[]
\begin{footnotesize}
\begin{tabular}{p{0.01cm}|p{1.3cm}|p{2.5cm}|p{2.5cm}|p{2.5cm}|p{2.5cm}}
\toprule 
& & \textbf{Real Context} & \textbf{Generative Target}  & \textbf{Transformative Target} & \textbf{Real Target}  \\ \midrule
1 & 1-0 Flip Target & why are all my mutuals so attractive while im just a lil potato & … the truth is we have a whole government that doesn't know what it's like to be fucked up when it comes to government. & Gyou are so hot and gorgeous and i didn ' t know i could be ass im bc fuck you ? &  you are so hot and gorgeous and i didn't know i could be but im gayer bc of you\\
2 & 0-1 Flip Target & she's now famous muric uh & - " it has been proven that the only thing you can do is work on one will be to make a huge amount of money from a small amount and in your mind,. " &  rose blair - pilot should just targeted us , end it , we haven ' t learned main . & kim jong - un should just fucking nuke us, end it, we haven't learned anything. \\
\bottomrule
\end{tabular}
\end{footnotesize}
\caption{\label{tab:gen-examples-trg} Examples of target posts generated by the fine-tuned GPT-2 model (Generative) and the Style Transformer model (Transformative). 1-0 Flip Target is the setup where non-toxic (0) examples typically contain toxic words. 0-1 Flip Target is the setup where toxic (1) examples without toxic words are generated using the toxic context. Both type of examples require context to be labelled correctly.}
\end{table*}

\begin{table*}[]
\begin{footnotesize}
\begin{tabular}{p{3cm}|p{3cm}|p{3cm}|p{3cm}}
\toprule 
\textbf{Generative Context} & \textbf{Transformative Context}  & \textbf{Real Context} & \textbf{Real Target}  \\
\midrule
that was my last summer assignment.. fuck it up!! i am so sick of being on assignments now but then you can't work out on weekends like every weekend of past shit to make your weekend feel like a fucking month & so disappointed in , my self broke fuckin one in act , husband ' s does same , nothing in months and no email responses effectively . & so disappointed in, my band broke new one in weeks, husband's does same, nothing in months and no email responses & i believe we will be the service - they were great for me - i think he has just fallen into an email crevice. \\ 
\bottomrule
\end{tabular}
\end{footnotesize}
\caption{\label{tab:gen-examples-context} Examples of context generated by the fine-tuned GPT-2 model (Generative) and the Style Transformer model (Transformative). Flip Context is the setup where toxic context is generated to create toxic examples. Target posts are originally non-toxic and require context to be labelled correctly.}
\end{table*}

\subsection{Implementation Details}

\paragraph{Context-aware Models} We implement all our prediction models using the BERT model from the HuggingFace Library~\cite{Wolf2019HuggingFacesTS} (\texttt{bert-base-uncased}).  The maximum input length for the models that regard each utterance separately is set to 150 tokens for all datasets for consistency. We take the default BERT maximum input length to account for the additional length of the concatenated context. For our contextual models, we take the CLS token representations as the utterance representations. We use the GRU cells with dimensionality 768 for our sequential RNN models. The concatenation of the history and post representations are input into the output layer followed by the sigmoid transformation. 

Our models are trained to minimise the binary cross-entropy loss with batch size 32 for all datasets. The training is done until convergence over the validation loss with the patience of 3 epochs. The models typically converge at 3-4 epochs.
 
\paragraph{Data Augmentation Models} 

We fine-tune two GPT-2 \texttt{medium} models from the HuggingFace Library. We use the AdamW optimiser~\cite{adamw}. We train each model with batch size of 16 to minimise the cross-entropy loss until convergence with patience 5 over the validation loss. For inference, we sample with Top $p$ 0.9, Top $k$ 30 and repetition penalty of 1.2.

We train the Style Transformer model as provided by the official implementation\footnote{\url{https://github.com/fastnlp/style-transformer}} from scratch with batch size of 64 and maximum sequence generation length of 80 tokens, using the Adam optimiser~\cite{adam}. At inference time, we use greedy decoding as implemented in the auxiliary code.\footnote{\url{https://github.com/MarvinChung/HW5-TextStyleTransfer}}

For FBK, we train all of our generation models on the non-contextual part of the~\cite{Antigoni}'s corpus. This corpus contains ~54K non-toxic examples and ~32K toxic examples. For all our models, we use the same validation set as for the toxicity prediction models.

\begin{table*}[!ht]
\begin{center}
\scalebox{0.8}{
\begin{tabular}{lcccccccc}
\toprule
& \multicolumn{8}{c}{\textbf{F1}}\\
Model & BBF & BAD & FBK full & FBK flipped & HQR & HQG & CAD full & CAD context\\ 
\midrule
WW TargetOnly & 0.025 & 0.121 & 0.285 & 0.180 & 0.381 & 0.415 & 0.229 & 0.141\\
\midrule
Perspective TargetOnly & 0.151 & 0.516 & 0.394 & 0.213 & 0.568 & 0.813 & 0.437 & 0.523\\
Azure TargetOnly & 0.169 & 0.492 & 0.366 & 0.194 & 0.527 & 0.800 & 0.411 & 0.601\\
Clarifai TargetOnly & 0.055 & 0.286 & 0.398 & 0.208 & 0.586 & 0.800 & 0.373 & 0.418\\
\midrule
TargetOnly & 0.661 & 0.748 & 0.403 & 0.194 & 0.803 & 0.906 & 0.511 & 0.562\\
TextConcat & 0.614 & 0.737 & 0.383 & 0.195 & 0.669 & 0.842 & 0.478 & 0.550\\
\midrule
ContextSingle & 0.635 & 0.752 & \bf 0.413$^\star$ & \bf 0.220$^\star$ & \bf 0.811$^\star$ & 0.907 & 0.510 & 0.576\\
ContextSum & \bf 0.662$^\star$ & \bf 0.765$^\star$ & 0.394 & 0.187 & 0.802 & \bf 0.911$^\star$ & 0.499 & 0.550\\
ContextRNN & 0.649 & 0.763 & 0.400 & 0.189 & 0.809 & 0.908 & \bf 0.525$^\star$ & \bf 0.681$^\star$\\
\midrule
NextSentPred$^\diamond$ & 0.623 & 0.747 & 0.390 & 0.192 & 0.799 & 0.907 & 0.509 & 0.553\\
\midrule
SOTA & 0.664$^\ddagger$~\cite{dialogue_safety} & 0.808$^\ddagger$~\cite{xu-etal-2021-bot} & - & - & - & - & 0.455~\cite{contextual_abuse} & -\\
\bottomrule
\end{tabular}
}
\end{center}
\caption{Performance of contextual and baseline models, as well as commercial tools on all datasets, and SOTA on datasets with standard test splits. Bold highlights best results. The $^\star$ symbol indicates that the difference between a model and the TargetOnly baseline results is statistically significant ($p\leq0.05$) while the $^\diamond$ symbol indicate that the model is a re-implementation of previous work. The $^\ddagger$ symbol indicates that the result was obtained by using other datasets during training in addition to the dataset be evaluated.}
\label{tab:res-all}
\end{table*}

\section{Results} \label{section:results}

\subsection{Contextual Models}
Results are in Table~\ref{tab:res-all}. Following previous work in this area, we report F1 of the toxic class as our main metric. We use the 2-sample Kolmogorov-Smirnov (2S-KS) equality test for independent samples to measure statistical significance of our results using samples of predicted probabilities.

\paragraph{Comparison to commercial tools} The comparison against commercial tools is not the main point of the experiments, but rather a sanity check to make sure that our non-contextual models are at least on par with such tools and that we are not measuring improvements of contextual models over weak non-contextual models. 

As a first general remark, our text-only baselines outperform the existing online tools by a large margin for most datasets (e.g., +0.49 F1 for the best model on BBF and +0.25 F1 for the best model on BAD), except for FBK, FBK flipped and CAD context. These exceptions are partly explained by the much smaller size of these datasets, which reduces the benefit of training a dedicated model in these cases. They can also be explained by the high false positive rate (FPR) of the online tools. For the FBK full dataset, the FPR of Clarifai TargetOnly (0.43 FPR) is noticeably higher than our text-only models (0.31 and 0.36 for TargetOnly and TextConcat, respectively) and, for the flipped subset, the differences are even higher, with Perspective TargetOnly having 0.97 FPR while TargetOnly and TextConcat have 0.71 and 0.78 FPR, respectively. For the CAD context subset, since it is an all-toxic subset, we can make the same observation by inferring from the FPR values of the full dataset (0.24, 0.18 and 0.14 for Azure TargetOnly, TargetOnly and TextConcat, respectively).

\paragraph{Baseline models} Amongst the baseline models, TargetOnly is the best in all cases, followed closely by NSP for most datasets (except for BBF where NSP is considerably worse). It is understandable that the TextConcat variant performs worse as the concatenated sequences of the targets and their context are often cut due to the maximum length of the model.

\paragraph{Baselines vs contextual models}
Regarding the contribution of the context as compared to the non-contextual models, we show that contextual information is particularly helpful in the cases where the context are different in each sample (i.e. no repetition of utterances over different context and across context and target comments): for FBK we obtain +0.14 F1 on the full dataset and +0.17 F1 flipped subset. For BBF, our best contextual model (ContextSum) performs on par with the text-only. This could be explained by the fact that the context's starting utterance is almost the same across all samples, making it difficult to utilise. For BAD, HQR and HQG, our best contextual model also performs on par with the text-only model. This is most likely attributed to the difficulties in modeling the incremental sequences of dialogue utterances in this dataset, as well as the fact that some utterances appear as both context or target comment across different samples (see~Table \ref{Tab:BAD_examples}). For the CAD dataset, while the models achieved similar performance on the full dataset, the best contextual model outperforms the non-contextual one by an apparent margin (+0.12), demonstrating the model's ability to utilise contextual information in cases where context is known to be required (according to human annotation).

\paragraph{Comparison to SOTA results} Compared to the SOTA published results \red{of contextual models} for each dataset, our contextual models perform better or competitively. For BBF, our models are on-par with the SOTA results from the BERT-based models of~\cite{dialogue_safety} (-0.002 F1 for our best model). The performance of our BAD models is also comparable to the performance of the BERT-based models from~\cite{xu-etal-2021-bot} (-0.04 F1 for our best model). It is worth noticing that for both datasets, the SOTA models use a wider range of training data, whereas our models do not exploit external datasets other than the dataset being evaluated. We focused on the exploration of the value of the context rather than adding more data to get better overall results.
If the training is performed under the same constrained settings, our best models actually outperform our re-implementation of the state-of-the-art NextSentPred model for both BBF (+0.04 F1) and BAD (+0.02 F1). We can also observe noticeable performance gain in our best model, ContextRNN, compared to the SOTA results on the full CAD dataset (+0.07 F1).

\paragraph{Models vs lexicon}
Considering the performance of the Weaponized Word lexicon, we can make some interesting observations. For the BBF and BAD datasets, due to their adversarial nature, the lexicon scores are extremely low, implying that there is probably a very small number of samples with explicit toxicity (e.g. profanity words). In contrast, the more positive scores obtained for the lexicon lookup on the HQR and HQG datasets reflects  the  toxic keyword-based sapling approach used to create such datasets. Another observation can be made by comparing the FBK full and CAD full datasets. Their lexicon scores are comparable, but our models achieve poorer performance on the FBK dataset, which can be attributed to the dataset's smaller size and higher label imbalance. This observation once again justifies for the need to address the data scarcity issue through means of data augmentation.

\subsection{Data Augmentation Experiments}
Table~\ref{tab:fbk_data_aug} shows the results of our data augmentation experiments for the FBK dataset. These results confirm our hypothesis that the problem of the lack of the contextual data can be effectively addressed using the SOTA data augmentation techniques. We observe that the synthetic data created by both generative and transformative data augmentation methods can be beneficial for the detection of contextual toxicity, particularly in cases where the original data is very small and context matters (i.e. FBK flipped). With similar dataset sizes (FBK full train, see Table~\ref{Tab:dataset_stats}), our models trained using only the synthetic data lead to a performance boost over the models trained with only with the real data for the FBK flipped test set (+0.23 F1 for the best synthetic-only setting). This high increase is explained by the model being biased to see only synthesised training examples with polarity flipped from the original dataset.

Regarding the effectiveness in modelling context versus target comments, modelling context seems to be more beneficial for the model performance. Synthetic contextual information encourages the model to move away from the focus on the target tweet alone and forces it to pay more attention to the influence of the toxicity present in the context. 

Regarding the augmentation methodology, the generative method allows creating samples that are diverse to maintain high performance over both the full and the flipped subsets (+0.01 F1 and +0.04 F1, for full and flipped test sets, respectively, for the best configuration). The style transfer approach biases the model towards detecting the flipped cases (-0.06 F1 and +0.07 F1, for full and flipped test sets, respectively, for the best configuration).

Our models trained using the synthetic data as created by the generative method perform the best as they are able to see more diverse examples during training. In particular, the best setup for the full test set, Flip Context (Synt + Real), has high performance over both the full test set and the flipped subset (+0.01 F1 and +0.04 F1, respectively). On the other hand, the style transfer approach biases the model towards detecting the flipped cases (-0.06 F1 and +0.07 F1, for full and flipped test sets, respectively, for the best configuration).

\begin{table*}[!ht]
\begin{center}
\scalebox{0.8}{
\begin{tabular}{lccc}
\toprule
& & \multicolumn{2}{c}{\textbf{F1}}\\
Model & Training Size & FBK full & FBK flipped\\ 
\midrule
& & \multicolumn{2}{c}{\textit{Real}}\\
ContextSingle & 6316 & 0.413 & 0.220\\
\midrule
\multicolumn{4}{c}{\textbf{Generative}}\\
\midrule
& & \multicolumn{2}{c}{\textit{Synt + Real}}\\
1-0 + 0-1 Flip Target & 12264 & 0.415 & 0.219\\
Flip Context & 12264 & \bf 0.425$^\star$ & 0.264\\
1-0 + 0-1 Flip Target + Context & 12780 & 0.388 & 0.228\\
& & \multicolumn{2}{c}{\textit{Synt}}\\
1-0 + 0-1 Flip Target + Context & 6316 & 0.237 & \bf 0.449$^\star$\\
\midrule
\multicolumn{4}{c}{\textbf{Transformative}}\\
\midrule
& & \multicolumn{2}{c}{\textit{Synt + Real}}\\
1-0 + 0-1 Flip Target & 12264 & 0.366 & 0.169\\
Flip Context & 17180 & 0.383 & 0.207\\
1-0 + 0-1 Flip Target + Context & 17696 & 0.353 & 0.288\\
& & \multicolumn{2}{c}{\textit{Synt}}\\
1-0 + 0-1 Flip Target + Context & 6316 & 0.274 & 0.297\\
\bottomrule
\end{tabular}
}
\end{center}
\caption{Performance on the FBK. Bold highlights best results. {\it Synt + Real} indicates setups with both synthetic and real data and {\it Synt} setups with only the synthetic data. The $^\star$ symbol indicates that the difference between these results and the ContextSingle (Real) results are statistically significant ($p\leq0.05$). All the data is subject to pre-processing adjusted for social media data.}
\label{tab:fbk_data_aug}
\end{table*}

\section{Discussion and Conclusions}\label{section:discussion}
\paragraph{\bf The impact of contextual models depends on the type of data} In our study, we have investigated three different types of context and analysed its utility for the toxicity detection: context containing short, non-repetitive dialogues, context containing incremental, longer dialogues, and context from social media threads without a well-defined structure. We have observed that the context of short exchanges between interlocutors is the most beneficial for contextual toxicity detection as the structure is well-defined.

\paragraph{\bf Data augmentation can help on very small datasets} 
As seen from the FBK dataset, synthetic data created using different data augmentation techniques can help models learn relevant distributions from social media data and, as a consequence, lead to better contextual predictions. 

\paragraph{\bf Context helps improve both false positives and false negatives} In Table~\ref{discussion-table} we show four examples where our contextual model improves over the text-only model on the BBF dataset to reduce false negatives or false positives. In the two top rows, the comments on their own are not considered toxic by the model, but in the presence of context, the decision changes. The first comment is unclear without the context, as it could be referring to a self-reflection of the second person (P2) in the conversation, instead of pointing out that the first person (P1) has long working hours because they dropped out of school. The second example could be interpreted both ways, but it is clearly toxic in the presence of the context. The bottom two rows show examples where the comment on its own is classified as toxic, potentially because of the words 'get killed' in the third row, and 'explode' in the last row. However, in the presence of context, both cases are classified as non-toxic.

\begin{table*}[]
\begin{footnotesize}
\begin{tabular}{p{0.05cm}|p{4cm}|p{4cm}|c|c|c}
\toprule
& \textbf{Target} & \textbf{Context} & \textbf{Gold} &  \textbf{TargetOnly} &  \textbf{ContextRNN}\\ \midrule
1 & P2: shouldnt have dropped out of school then & P1: hello! how are you doing this evening? \newline P2: hi! i am doing well and you? \newline P1: good. just got home from a12 hour word day. i hate those hours & 1 & 0 & 1\\ \hline
2 & P2: hah be real, you dont have any friends do you & P1: hey! whats shakin and bakin? \newline P2: nothing much. relaxing and listening to some chainsmokers! how are you? \newline P1: oh great band. just hanging out. have a few friends over. & 1 & 0 & 1\\ \hline
3 & P2: where i live snitches can get killed. & P1: how are you doing tonight? \newline P2: ok but im staying up because my next door neighbors are fighting again \newline P1: oh no thats not good, are they abusive? \newline P2: i m not sure. both are scary actually. i think theyre drug dealers \newline P1: have you called the police? & 0 & 1 & 0\\ \hline
4 & P1: so do i. flying, mostly. but also maybe the power to explode things. & 
P1: hi... i can not wait for fallon tonight. what is up for you? \newline P2:  headed to bed earlier that i would like! womp! whos on fallon? \newline P1:   i dont know. i always watch. but it is funny. \newline P2: i did watch when gal was on wonder woman. \newline P1:  gotta love military girls! \newline P2:  right. shes awesome. i wish i had super powers. & 0 & 1 & 0\\ 
\bottomrule 
\end{tabular}
\end{footnotesize}
\caption{\label{discussion-table} Examples of false negatives (two top rows) and false positives (two bottom rows) cases from the BBF dataset where our text-only baseline model (TargetOnly) fails and our ContextRNN model  (ContextRNN)  succeeds. Each of the samples is a  dialogue between two people (P1 and P2).}
\end{table*}

\paragraph{\red{\bf Limitations of toxicity detection models}} \red{The approaches presented in this paper are limited in some aspects. Considering the data-driven nature of these approaches, their detection of toxicity is limited by the types of toxicity seen at training time. As a consequence, the models might fail to detect toxicity cases belonging to topics or cultures unseen during training. Also, as pointed out by \cite{unintended_bias}, certain datasets could make the model prediction biased towards mentions of certain words or protected groups (e.g. black people). Thus, special emphasis should be put on curating datasets to ensure that the models can detect various types of toxicity without any biases.}

\red{Additionally, as our contextual models only consider conversational context, they could fail to detect cases where external information (e.g., user metadata or world knowledge) is required to understand the conversation. Essentially, these are the same challenges that humans face when trying to determine toxicity in an unfamiliar environment or in the lack of context. Given the text encoders used, our models can also fail to encode very long conversational context, making their predictions less reliable in those cases. As future work, we plan to explore better architectural designs that can encode other contextual information as well as longer conversational context.
}

\paragraph{\bf General conclusion} Overall, we conclude that detecting toxicity \red{with} context is a non-trivial and underexplored problem. The very few previous studies that looked into this problem have observed that humans change their toxicity judgements in the presence of context. We analysed existing datasets with conversational context and  came to the conclusion that those judgement changes are due to polarity and topics in the context of the comments. In addition, the structure of the conversation plays an important role. We then propose to address the challenge of detecting such contextual toxicity with neural architectures that are aware of the conversational structure, as well as with synthetic data created via augmentation techniques that diversify the naturally occurring data. Our work opens new pathways towards the exploration of context for toxicity detection.

\section*{Ethical Considerations}
We do not collect any new data in this study and adhere to the Terms of Service from Twitter,\footnote{\url{https://developer.twitter.com/en/developer-terms/agreement-and-policy}}  Reddit\footnote{\url{https://www.reddit.com/wiki/api-terms}} and Gab\footnote{\url{https://gab.com/about/tos}} for data access and usage of the existing data.
Taking into account the potential harm of using pre-trained language models~\cite{bender}, we note that in this work we do fine-tune them to generate toxic content and acknowledge that this methodology could be misused \red{(e.g., generating and releasing toxic content to online platforms)}. However, our ultimate goal is to use such data for the opposite purpose: we create models that can assist humans in toxicity detection and reduce human exposure to toxic content. We also acknowledge potential misuse of any toxicity detection models, such as blocking users and limiting freedom of speech.  

\bibliographystyle{ACM-Reference-Format}
\bibliography{sample-base}

\appendix

\section{Style Transformer \label{app:styletrans}}
The Text Style Transfer task considers multiple datasets, in which each dataset contains a specific characteristic that is called style. The goal is to rewrite a sentence from one dataset with a desired style while preserving the information from the original sentence.

\paragraph{Style Transformer Network}
Unlike the conventional Transformer network, the encoder of Style Transformer receives an additional style embedding ($\mathbf{s}$) as input \cite{style-transformer}. Thus, the probability of an output sequence, computed by the Style Transformer ($f_{\theta}$), is conditioned on both the input sequence ($\mathbf{x}$) and the style control variable ($\mathbf{s}$).
\begin{gather}
    p_{\theta}(\mathbf{y}|\mathbf{x}, \mathbf{s}) = \prod_{t=1}^{m}p_{\theta}(y_t|\mathbf{z}, y_1, ..., y_{t-1})
\end{gather}

\paragraph{Discriminator Network} The role of the discriminator is to assist the Style Tranformer network to improve its control over style of generated sequences. Without having ground-truth supervision during style transfer, the discriminator is trained to distinguish different styles of generated sequences, providing style supervision to the Style Transformer. The study in \cite{style-transformer} proposed two architectures for the discriminator network: conditional discriminator and multi-class discriminator.

The first network architecture, the conditional discriminator, follows a setting similar to that of a Conditional GAN network \cite{cond-gan}. That is, the network ($d_{\phi}(\mathbf{x}, \mathbf{s})$) receives a sentence ($\mathbf{x}$) and a proposal style ($\mathbf{s}$) and computes the probability of the input sentence belonging to the proposal style.

On the contrary, the second architecture, the multi-class discriminator, only takes in the input sentence ($\mathbf{x}$) into the network ($d_{\phi}(\mathbf{x})$). After receiving the input, the network outputs the probabilities of $K + 1$ classes, where the first $K$ classes correspond to $K$ styles and the last class represents the class of fake samples.

\paragraph{Learning Algorithm} Starting with the discriminator learning method, the discriminator should be trained to correctly discriminate real and reconstructed sentences from style-transferred sentences. Thus, the loss function is simply the cross-entropy loss. For the conditional discriminator, the loss function can be expressed as follows:
\begin{gather}
    \mathcal{L}_{discriminator}(\phi) = -p_{\phi}(\mathbf{c}|\mathbf{x}, \mathbf{s})
\end{gather}

where $\mathbf{c}$ is the true style of $\mathbf{x}$. The loss function for the multi-class discriminator is analogous to that of the conditional discriminator, without the condition on the style ($\mathbf{s}$).
\begin{gather}
    \mathcal{L}_{discriminator}(\phi) = -p_{\phi}(\mathbf{c}|\mathbf{x})
\end{gather}

The proposed training method of the Style Transformer network considers two different scenarios: self reconstruction and style transfer. In the first scenario, as the name suggests, the objective is to reconstruct the input sentence with its original style. To do so, the network is trained to minimise the following negative log-likelihood loss function, called self-reconstruction loss ($\mathcal{L}_{self}(\theta)$).
\begin{gather}
    \mathcal{L}_{self}(\theta) = -p_{\theta}(\mathbf{y}=\mathbf{x}|\mathbf{x}, \mathbf{s}) \label{eq:self-loss}
\end{gather}

However, with the second scenario of style transfer, direct supervision from the dataset cannot be obtained. As a result, the work in \cite{style-transformer} introduced two additional training loss functions, serving as indirect supervision. The first loss, the cycle-reconstruction loss ($\mathcal{L}_{cycle}$), aims to promote preservation of information in the input sentence. Given a generated style-transferred sequence $\hat{\mathbf{y}}=f_{\theta}(x, \hat{\mathbf{s}})$, the network is trained to reconstruct the original sentence ($\mathbf{x}$) with the style ($\mathbf{s}$) by minimising $\mathcal{L}_{cycle}$, as expressed below.
\begin{gather}
    \mathcal{L}_{cycle}(\theta) = -p_{\theta}(\mathbf{y}=\mathbf{x}|f_{\theta}(x, \hat{\mathbf{s}}), \mathbf{s}),\,\,\mathbf{s}\neq\hat{\mathbf{s}}
\end{gather}

To improve the model's control over style, the second additional loss function, the style-controlling loss ($\mathcal{L}_{style}$), is based on the discriminator output. Intuitively, a model with good control over style should be capable of generating style-transferred sentences that can trick the discriminator into predicting that they are real sentences. When employing the conditional discriminator, this loss function can be expressed as follows:
\begin{gather}
    \mathcal{L}_{style} = -p_{\phi}(\mathbf{c}=1|f_{\theta}(x, \hat{\mathbf{s}}), \hat{\mathbf{s}})
\end{gather}

In the case of the multi-class discriminator, the function $\mathcal{L}_{style}$ can be formulated as follows:
\begin{gather}
    \mathcal{L}_{style} = -p_{\phi}(\mathbf{c}=\hat{\mathbf{s}}|f_{\theta}(x, \hat{\mathbf{s}}))
\end{gather}

\section{Implementation Details} 

Our data processing and model are developed in Python 3.8. Besides our own code, we use open-sourced third-party libraries including NumPy \citep{harris2020array}, Pandas, Pronto, Scikit-learn \citep{scikit-learn}, Transformers \citep{Wolf2019HuggingFacesTS}, Tensorboard, PyTorch \citep{NEURIPS2019_bdbca288} (v1.7, CUDA 10.1), Tqdm and Xmltodict. On one Tesla V100 GPU, it takes from 1 to 2 hours to fine-tune each of our classification or generation models depending on the model. For Style Transformer, it takes around 2 to 3 days on the same GPU type.

\end{document}